\documentclass[10pt,twocolumn,letterpaper]{article}

\usepackage{cvpr}              

\usepackage{algorithm}
\usepackage{algpseudocode}
\usepackage{makecell} 
\usepackage{times}
\usepackage{latexsym}
\usepackage{amsmath}
\usepackage[utf8]{inputenc}
\DeclareUnicodeCharacter{202A}{}
\usepackage{graphicx}

\definecolor{cvprblue}{rgb}{0.21,0.49,0.74}
\usepackage[pagebackref,breaklinks,colorlinks,allcolors=cvprblue]{hyperref}

\def\paperID{*****} 
\def\confName{CVPR}
\def\confYear{2026}

\title{Lightweight and Production-Ready PDF Visual Element Parsing}

\author{Meizhu Liu \quad Yassi Abbasi \quad Matthew Rowe \quad Michael Avendi \quad  Paul Li \\ 
 Oracle AI}

\begin{document}
\maketitle

\begin{abstract}
PDF documents contain critical visual elements such as figures, tables, and forms whose accurate extraction is essential for document understanding and multimodal retrieval-augmented generation (RAG). Existing PDF parsers often miss complex visuals, extract non-informative artifacts (e.g., watermarks, logos), produce fragmented elements, and fail to reliably associate captions with their corresponding elements, which degrades downstream retrieval and question answering. We present a lightweight and production level PDF parsing framework that can accurately detect visual elements and associates captions using a combination of spatial heuristics, layout analysis, and semantic similarity. On popular benchmark datasets and internal product data, the proposed  solution achieves $\geq96\%$ visual element detection accuracy and $93\%$ caption association accuracy. When used as a preprocessing step for multimodal RAG, it significantly outperforms state-of-the-art parsers and large vision-language models on both internal data and the MMDocRAG benchmark, while reducing latency by over $2\times$. We have deployed the proposed system in challenging production environment. 
\end{abstract}

\maketitle

\section{Introduction}
Accurate understanding of visual elements from PDF documents is a foundational requirement for document analysis, automation, and multimodal retrieval-augmented generation (RAG). Modern PDFs interleave text with tables, figures, forms, and diagrams in highly heterogeneous layouts, making reliable parsing challenging. Errors at this stage—such as missing tables, fragmented images, or incorrect caption alignment—propagate and substantially degrade downstream retrieval and question answering performance.

Prior work on PDF parsing can be broadly grouped into heuristic-based tools (e.g., PDFPlumber \cite{pdfplumber}, PyMuPDF \cite{pymupdf}, PDFMiner \cite{pdfminer}, Apache Tika \cite{tika}), layout learning approaches (e.g., PubLayNet \cite{zhong2019publaynet}, DocBank \cite{li2020docbank}, DocLayout-YOLO \cite{DOCLAYOUT-YOLO} , Unstructured \cite{unstructured} ), and vision-language models. Heuristic tools provide efficient access to low-level PDF primitives but struggle with complex layouts, forms, and caption alignment. Layout learning methods improve structural detection but often lack robustness across domains and may not address visual noise or caption association. Recent VLMs demonstrate strong end-to-end PDF understanding ability  but incur high computational cost and latency and are unsuitable for large-scale deployment. Our work complements these approaches by combining lightweight heuristics with semantic reasoning, achieving competitive accuracy with substantially lower latency. Our major contributions could be summarized as follows. 

\paragraph{Table detection with spatial and graphical heuristics.}

We improve table identification accuracy by combining graphical analysis with spatial alignment heuristics. 
Bordered tables are detected through vector graphics analysis (e.g., counting straight lines), while unbordered tables are identified by detecting repeated text block alignments into rows and columns. 
This approach effectively handles both explicitly structured and visually implicit table layouts without requiring predefined templates.

\paragraph{Layout-driven form detection via text block profiling.}

We introduce an adaptive form detection method that profiles text blocks based on density and character length. 
Pages with a high occurrence of short snippets—typical of form fields and labels—are automatically flagged, removing the need for predefined templates or field dictionaries.

\paragraph{Redundancy-resilient image merging.}

We propose a post-processing mechanism to merge overlapping or duplicate image extractions by comparing bounding box coordinates and applying overlap thresholds. 
This ensures that only a single, accurate instance of each visual element is preserved, which is crucial for technical diagrams and blueprints where fragmentation can distort meaning.

\paragraph{Multi-tiered visual noise filtering.}

We develop a robust filtering strategy to remove non-essential visual artifacts (e.g., watermarks, logos) by combining: 1) spatial analysis such as position and size, 2)  appearance frequency across pages, 3) transparency detection via alpha channels, 4) semantic similarity via embedding vectors.
This combination of heuristics and semantic reasoning substantially improves precision in distinguishing useful content from noise.

\paragraph{Caption extraction with semantic-layout fusion.}
We design a hybrid method to associate captions with visuals from the following angles: 1) geometric relationships between visuals and nearby text, 2) spatial clustering of text blocks, 3) stylistic cues including font size and boldness, 4) semantic embedding similarity.
This enables accurate title and caption extraction even in poorly structured or inconsistently formatted documents.


\paragraph{An efficient, production-ready PDF parsing framework.}
We integrate all components into a unified, lightweight PDF parsing system that enables accurate detection and semantic association of visual elements at scale. The resulting document representations are cleaner and more structured, substantially benefiting downstream applications such as multimodal retrieval and document question answering.


\section{Proposed Method}

We present a production-ready unified framework for detecting and extracting key visual elements—images, tables, and forms—from PDF documents, while robustly associating each element with its corresponding caption or title under complex and inconsistent layouts.


The system consists of four main components:
\begin{enumerate}
\item heuristic-based table and form detection,
\item detection and consolidation of duplicate or fragmented images,
\item filtering of non-informative artifacts (e.g., logos and watermarks), and
\item caption association using combined layout and semantic cues.
\end{enumerate}
 
Our pipeline leverages \texttt{PyMuPDF} to parse each PDF page and extract three modalities: embedded images with bounding boxes, text blocks, and low-level drawing primitives (e.g., lines and rectangles). On top of this raw output, we apply heuristic algorithms to infer higher-level structures. The extraction pipeline includes the following modules, each of which will be presented in the coming sections.

\subsection{Form Detection}
\label{sec:form}

\begin{itemize}
\item \textbf{Images:} Detected via \texttt{page.get\_images} and extracted using \texttt{doc.extract\_image}.
\item \textbf{Bounding boxes:} Obtained from \texttt{page.get\_image\_rects}.
\item \textbf{Text blocks:} Retrieved using \texttt{page.get\_text("blocks")} with positional metadata.
\item \textbf{Tables:} Identified through spatial heuristics over text block alignment and density.
\item \textbf{Forms:} Detected using layout- and widget-based visual heuristics.
\item \textbf{Captions and titles:} Matched to visual elements using proximity-based layout cues combined with semantic similarity.
\item \textbf{Artifact filtering:} Discriminative rules separate meaningful figures from decorative or repetitive visuals.
\end{itemize}

\begin{algorithm}[h]
\caption{Detect Table in PDF Page}
\begin{algorithmic}[1]
\Procedure{PageContainsTable}{$\text{page}$}
    \If{$|\{d \in \text{page.get\_drawings()} : d.\text{type}=\text{"line"}\}| \ge K$}
        \State \Return \textbf{True} \Comment{Bordered table}
    \EndIf

    \State $\text{blocks} \gets \{(x,y): (x,y,t)\in\text{page.get\_text("blocks")},\ t\neq\emptyset\}$
    \State $\text{x\_cnt} \gets \text{Count}(\text{round}(x)),\ 
           \text{y\_cnt} \gets \text{Count}(\text{round}(y))$

    \If{$|\{v\in\text{x\_cnt}: v\ge3\}|\ge M$ \textbf{and} 
        $|\{v\in\text{y\_cnt}: v\ge3\}|\ge M$}
        \State \Return \textbf{True} \Comment{Unbordered table}
    \EndIf
    \State \Return \textbf{False}
\EndProcedure
\end{algorithmic}
\label{tabledetection}
\end{algorithm}

\subsection{Table Extraction}
\label{sec:table}

We improve table detection using spatial heuristics that exploit text alignment and graphical structure (Algorithm~\ref{tabledetection}), and it could provide layout signals robust to font variation, language, and OCR noise. Using PyMuPDF primitives, the method detects both bordered and unbordered tables. For bordered tables, vector drawings are analyzed to count straight lines; pages with at least $K$ lines are classified as containing tables. For unbordered tables, we examine text block coordinates, where repeated and aligned $x$ and $y$ positions reveal column and row structures. Pages exhibiting at least $M$ distinct vertical and horizontal alignments are classified as containing tables, reducing false negatives and improving robustness across diverse PDF layouts.

\begin{algorithm}
\caption{Detection of AcroForm Widgets in PDF Pages}
\label{alg:acroformdetection}
\begin{algorithmic}[1]
\Require PDF document $D$
\Ensure List of detected form fields and their metadata
\ForAll{page $p \in D$}
    \State \texttt{widgets} $\gets$ \texttt{p.widgets()}
    \ForAll{widget $w \in$ \texttt{widgets}}
        \State \texttt{field\_name} $\gets$ $w.\texttt{field\_name}$
        \State \texttt{field\_type} $\gets$ $w.\texttt{field\_type}$
        \State \texttt{rect} $\gets$ $w.\texttt{rect}$
        \State \textbf{output} (\texttt{field\_name}, \texttt{field\_type}, \texttt{rect})
    \EndFor
\EndFor
\end{algorithmic}
\end{algorithm}

Since PyMuPDF does not natively support form detection, we identify form-like content using three complementary strategies:
\begin{itemize}
\item \textbf{Explicit PDF form fields (AcroForms):} For PDFs with embedded form fields, we extract interactive elements (e.g., text inputs, checkboxes) via the AcroForm  \cite{pdf32000} using \texttt{page.widgets()} \cite{pymupdf_docs}, including field types and bounding boxes. The whole pipeline can found in  Agorithm \ref{alg:acroformdetection}.
\item \textbf{Visually structured forms without fields:} For forms defined purely by layout, we analyze text blocks and treat short text spans (text blocks containing only a small number of characters or words, e.g., “Name”, “Date”, “Email”) as potential field labels. Pages containing a high density of such short spans are classified as form-like. Pages with more than $T_b$ blocks shorter than $T_t$ are classified as form-like, which is effective for printed or scanned forms.
\item \textbf{Structural drawing elements:} We extract line and rectangle primitives using \texttt{page.get\_drawings()} to capture boxes and separators commonly used in form layouts.
\end{itemize}




\subsection{Handling Duplicate and Fragmented Images}
\label{sec:dedup}
To address duplicate and fragmented image extraction—such as overlapping subimages or segmented parts of a single image—we apply a post-processing step based on spatial overlap. Bounding boxes and dimensions of all extracted images are compared, and images with overlap exceeding a predefined threshold $\theta$ are treated as redundant or belonging to the same visual element. These regions are then merged into a single unified image to remove duplication and improve accuracy. 
 To be specific, suppose two images $Img_i$ and $Img_j$ are detected, and 

\begin{equation}
\label{eqn:imageduplication}
\frac{|Img_i \cap Img_j|}{|Img_i| \cup |Img_j|} > \theta,
\end{equation}
where $\theta$ is a hyperparameter which can be tuned on the training dataset. 
If Eqn. (\ref{eqn:imageduplication}) holds, then we merge $Img_i$ and $Img_j$ into one. In this way, we were able to remove duplicated images, and avoid fragmented images. 





\subsection{Distinguishing Meaningful Visual Content From Non-essential Artifacts}
\label{sec:filter}

One of the critical challenges in PDF document parsing is differentiating between meaningful visual content (e.g., figures, charts) and non-essential artifacts such as watermarks and logos. Since tools like PyMuPDF treat all embedded visuals uniformly, PyMuPDF will extract the logo and the watermark), downstream processing must filter out these noisy elements to ensure clean data extraction.


Watermarks and logos are embedded similarly to genuine illustrations, making them difficult to distinguish without heuristic filtering or post-processing. To address this, we introduce a set of rule-based algorithms that filter out non-essential images using three core criteria: size, position, and frequency.

These non-informative elements often exhibit common visual patterns:
\begin{itemize}
   \item \hspace{2em}  Repeated on most or all pages of the document
    \item \hspace{2em} Fixed or small in size
    \item  \hspace{2em} Frequently semi-transparent
    \item  \hspace{2em} Consistently positioned (e.g., corners, center diagonals)
\end{itemize}

Our proposed filtering strategy includes:

\begin{itemize}
    \item \textbf{Size-based filtering:} Discards images smaller than a threshold ($p$\% of the page width or height, or if width $<N$ and height $<N$). These thresholds are empirically derived from the production dataset statistics.
    
    \item \textbf{Position- and embedding-based filtering:} Images located in high-likelihood zones (e.g., corners) are compared against a logo database using CLIP-based embeddings. Details are in Section~\ref{subsub:logorecognition}.
    
    \item \textbf{Frequency-based filtering:} Images that appear on over $q$\% of pages (in documents with more than one page) are flagged as likely artifacts.
    
    \item \textbf{Alpha channel filtering:} Semi-transparent visuals are detected using alpha channel inspection. These images often indicate watermarks.
\end{itemize}

\subsubsection{Logo Recognition}
\label{subsub:logorecognition}

To recognize logos, we analyze images located in regions typically reserved for branding (e.g., top-left, bottom-right corners). Each image is passed through OpenAI’s CLIP encoder to generate an embedding vector. We then compute the cosine similarity between the image embedding and a reference logo dataset (e.g., Logo2K~\cite{Wang2020Logo2K}). If the similarity exceeds a threshold $\delta$, the image is classified as a logo and removed.

\subsubsection{Watermark Detection}
\label{subsub:watermark}

Watermark identification leverages a combination of alpha channel analysis, grayscale thresholding, connected components filtering, and OCR.

First, we examine the alpha channel to detect transparency. An image is marked as semi-transparent if the minimum pixel value in the alpha channel is less than 255. More specifically:
\begin{itemize}
    \item We extract the region's alpha mask.
    \item Compute average opacity.
    \item Discard if opacity is below threshold $\tau$.
\end{itemize}

Additionally, we convert images to grayscale and apply adaptive or OTSU thresholding to isolate low-contrast watermark patterns. We use connected component labeling to isolate distinct elements and remove regions that are too large, small, or noisy.

To confirm watermark identity, OCR (e.g., Tesseract) is applied to extract embedded text. Detected strings are matched against a predefined vocabulary of watermark keywords:
\begin{quote}
``Confidential'', ``Do Not Copy'', ``Sample'', ``Draft'', ``Internal Use Only'', ``Restricted'', ``Top Secret''
\end{quote}

The watermark detection algorithm is detailed in Algorithm~\ref{alg:watermark}.

\begin{algorithm}
\caption{Watermark Detection via Transparency and OCR}
\label{alg:watermark}
\begin{algorithmic}[1]
\Require PDF image $I$
\Ensure Boolean indicating whether $I$ is a watermark
\If{$I$ has alpha channel}
    \State $\alpha \gets$ extract alpha channel
    \If{$\min(\alpha) < 255$}
        \State Compute average opacity per region
        \If{opacity $< \tau$}
            \State mark region as semi-transparent
        \EndIf
    \EndIf
\EndIf
\State Convert $I$ to grayscale and apply thresholding
\State Detect components and discard noise
\State Apply OCR on $I$ to extract text $T$
\If{$T$ contains known watermark keywords}
    \State \Return True
\Else
    \State \Return False
\EndIf
\end{algorithmic}
\end{algorithm}

\begin{algorithm}[h]
\caption{Caption Extraction via Semantic-Layout Fusion}
\label{alg:caption_extraction}
\begin{algorithmic}[1]
\Require PDF page $p$, visual elements $V = \{v_1, v_2, \dots\}$, fusion weight $\alpha$
\Ensure For each $v_i \in V$, return best-matched caption $C_i$

\ForAll{visual element $v_i \in V$}
  \State $B_i \gets$ bounding box of $v_i$ (using PyMuPDF)
  \State $W_i \gets$ vertical context window around $B_i$ (scaled to height of $B_i$)
  \State $T_i \gets$ all text lines within $W_i$
  \State $B_i^{text} \gets$ cluster $T_i$ into blocks using DBSCAN

  \ForAll{text block $b_j \in B_i^{text}$}
    \State $H_j \gets$ layout-based score of $b_j$ using:
      \begin{itemize}
        \item keyword presence (e.g., ``Figure'', ``Table'')
        \item distance to $B_i$
        \item font size and style (bold/italic)
        \item enumeration pattern (e.g., ``Fig. 2'')
      \end{itemize}
    \State $S_j \gets$ semantic similarity score between $b_j$ and $v_i$ (e.g., via CLIP)
    \State $F_j \gets \alpha \cdot H_j + (1 - \alpha) \cdot S_j$
  \EndFor

  \State $C_i \gets b_j$ with highest $F_j$
\EndFor

\State \Return $\{(v_i, C_i)\}$ for all visual elements
\end{algorithmic}
\end{algorithm}

\subsection{Caption Extraction via Semantic-Layout Fusion}
\label{sec:caption}

Identifying captions or titles for images, tables, charts, and forms in PDF documents is particularly challenging due to inconsistent formatting and ambiguous layout structures. We propose a fusion-based method that integrates both visual layout heuristics and semantic similarity modeling to robustly associate visual elements with their corresponding captions or titles. The proposed approach consists of the following key components, and the overall procedure is outlined in Algorithm~\ref{alg:caption_extraction}:

\begin{itemize}
    \item \textbf{Visual Anchor Detection:} Visual elements such as images, tables, and forms are detected using PyMuPDF, and their bounding boxes are extracted to serve as anchor regions for subsequent caption search.

    \item \textbf{Context Window Construction:} For each visual anchor, a vertical context window  is constructed above and below its bounding box, including the current page as well as the preceding and following pages. The window size is adaptively scaled based on the height of the visual element to capture relevant surrounding text.

    \item \textbf{Text Line Clustering and Filtering:} Text lines falling within the context window are extracted and grouped into coherent text blocks using a density-based clustering algorithm DBSCAN (with parameter $\epsilon$). This step aids in distinguishing isolated titles from paragraph-like content.

    \item \textbf{Layout-Based Prioritization:} Candidate text blocks are ranked using visual layout features, including font size, boldness, italics, and alignment. Such features often signal titles or captions due to their distinct formatting compared to body text.

    \item \textbf{Heuristic Title Scoring:} Each candidate block receives a layout-based heuristic score computed based on:
    \begin{itemize}
        \item Presence of caption-specific keywords (e.g., \textit{Figure}, \textit{Fig.}, \textit{Table}, \textit{Chart}, \textit{Diagram}, \textit{Form})
        \item Spatial proximity to the visual element
        \item Font styling indicators (e.g., larger size, boldness)
        \item Enumeration patterns (e.g., ``Fig. 2'', ``Table 3'')
    \end{itemize}

    \item \textbf{Semantic Title Scoring:} To complement layout-based analysis, we compute a semantic similarity score between each candidate block and its associated visual element. This is achieved using a multimodal embedding model (e.g., CLIP). Candidate blocks that are semantically aligned with the visual content are preferred, particularly when layout cues are ambiguous.
\end{itemize}

 \subsection{Text Extraction and Postprocessing}
\label{sec:textextraction}

PDFs may encode letter ligatures (e.g., \texttt{U+FB01}) and non-standard whitespace, which can introduce artifacts during text extraction. We normalize ligatures to their plain equivalents, replace non-breaking and invisible spaces with standard spaces, and rejoin words split across line breaks by removing spurious hyphenation.

\begin{table*}[ht]
\centering
\setlength{\tabcolsep}{2pt}
\begin{tabular}{|l|c|c|c|c|c|c|c|c|c|}
\hline
\textbf{Tool} & \textbf{{Text}} & \textbf{ {Cap. sim}} & \textbf{ {Tbl BBA}}  & \textbf{ {Img BBA}} & \textbf{ {Frm BBA}} & \textbf{DC} & \textbf{ {Tbl DC}} & \textbf{ {Img DC}} & \textbf{ {Frm DC}}   \\
\hline
\texttt{PDFPlumber}  &0.97 & X   & 81\%    &     X      &    X         &  88\%&  88\% &  X&  X \\
\texttt{PyMuPDF}      &0.98 & 0.75        & 76\%     & 95\%    & X    & 82\% & 79\% & 91\% & X  \\
\texttt{Tika}         &0.98 & X        & X              & 82\%      & X             & 69\%   & X & 69\% & X   \\
\texttt{{unstructured}} & 0.97 & X     &  82\%          & 86\%  & X   & 73\% & 85\%& 72\% & X \\
\texttt{PDFMiner}     &0.97 & X & X &  96\%       & X    & 91\% &X &91\% &X \\
\texttt{Proposed}    & \textbf{0.99} & \textbf{0.93}  & \textbf{96\%}    & \textbf{98\%}    &  \textbf{100\%}  &  \textbf{100\%}&  \textbf{100\%}&  \textbf{100\%}&  \textbf{100\%}   \\
\hline
\end{tabular}
\caption{Performance comparison of PDF parsing tools across text extraction accuracy, caption similarity, table detection bounding box accuracy (BBA), image detection BBA, form detection BBA, overall detection completeness (DC), table DC, image DC and form DC. "X" means the tool does not support the corresponding functionality.}
\label{tab:pdf_parsing}
\end{table*}

\begin{table*}[ht]
\centering
\begin{tabular}{|l|c|c|c|c|c|c|c|}
\hline
\textbf{Model} & \textbf{Text} & \textbf{Cap. sim} & \textbf{Tbl BBA}  & \textbf{Img BBA} & \textbf{Frm BBA} & \textbf{DC} &\textbf{Latency (s)}  \\
\hline
\texttt{GPT4.1}       &0.99 & 0.89  & 86\% & 96\% & 97\%  & 87\% & 10.7 \\
\texttt{GPT4.1-mini}  &0.98 & 0.85  & 84\% & 95\% & 81\%  & 82\% &  8.2 \\
 \texttt{Gemini2.5Pro}&0.98 & 0.84  & 82\% & 93\% & 95\%  & 64\% & 14.3 \\
\texttt{Proposed}     &\textbf{0.99} & \textbf{0.93}  & \textbf{96\%} & \textbf{98\%} &\textbf{100\%}  &\textbf{100\%} & \textbf{4.1} \\
\hline
\end{tabular}
\caption{Performance comparison of PDF parsing with different VLMs. The evaluation metrics are Text accuracy, Caption similarity, Table BBA (table detection bounding box accuracy), Image BBA, Form  BBA, and DC (detection completeness), and latency in terms of the mean time (s) per page.}
\label{tab:comparewithVLM}
\end{table*}

 %

\section{Experimental Results}



 We evaluate our method on internal product data and multiple public benchmarks, including MMDocRAG \cite{dong2025benchmarkingretrievalaugmentedmultimomalgeneration}, PDFVQA \cite{pdfvqa,pdfvqa2}, DocVQA \cite{docvqa}, and PDF-MVQA \cite{PDFMVQA}. Parameters are tuned on the MMDocRAG dev split via Optuna \cite{optuna} and reported in Table~\ref{tab:parameters}. The evaluation covers text extraction, visual element detection (images, tables, forms), caption association, and downstream document question answering \cite{liu2013, liu2016}. 

\begin{table}[ht]
\centering
\scriptsize
\setlength{\tabcolsep}{5.5pt}
\begin{tabular}{|l|c|c|c|c|c|c|c|c|c|c|c|}
\hline
\textbf{$K$} &\textbf{$M$} & \textbf{$T_b$/$T_t$} & \textbf{$\theta$} & \textbf{$p$} & \textbf{$q$}  & \textbf{$N$}& $\delta$ &  \textbf{$\tau$} & \textbf{$\alpha$} & \textbf{$\epsilon$} \\
\hline   5 & 2 & 30/5   &  0.25 & 5 & 80 & 50 & 0.75 & 0.8     & 0.45   & 1.5 \\
\hline
\end{tabular}
\caption{Main Hyper-Parameter used in our experiments.}
\label{tab:parameters}
\end{table}

Our internal production dataset consists of 21 documents totaling approximately 400 pages. These pages contain a diverse mix of content, including text, images (176), tables (142), forms (4), watermarks (146), and logos (293). 6 PDF files are scanned documents and the rest are vector PDFs. The dataset spans multiple domains, including user manuals, guidebooks, research papers, news, W8 and W9 forms. The dataset has complex layouts, e.g. there are multiple figures or tables or charts next to each other or one below the other. Each page is annotated with expert-labeled ground truth texts and visual elements, including element bounding boxes and associated captions.
We extracted text, visual elements and identified the captions for the visual elements, from each page using our proposed pipeline. We evaluated the performance using the following metrics:
 
\begin{itemize}
    \item \textbf{Text Extraction Accuracy:}  
    We evaluate the textual similarity between the extracted text $T_p$ and the ground truth $T_{gt}$ using the Levenshtein  Distance \cite{levenshtein1966binary}. The similarity is defined as:

    \begin{equation} 
    \label{eqn:sim}
    \text{Similarity}(T_p, T_{gt}) = 1 - \frac{\text{Levenshtein}(T_p, T_{gt})}{\max(\lvert T_p \rvert, \lvert T_{gt} \rvert)}.
    \end{equation}

    \item \textbf{Bounding Box Accuracy (BBA):}  
    To assess the accuracy of detecting images, tables and forms, we compute the Intersection over Union (IoU) between predicted bounding boxes $B_p$ and ground truth boxes $B_{gt}$:

    \begin{equation} 
    \text{IoU}(B_p, B_{gt}) = \frac{|B_p \cap B_{gt}|}{|B_p \cup B_{gt}|}.
    \end{equation}

    A predicted box is considered correct if $\text{IoU} \geq \tau$, with $\tau$ adjustable based on the datasets (we set $\tau$ to 0.8 in our experiments). The overall BBA accuracy is:

    \begin{equation}
    \text{BBA} = \frac{\text{Number of correctly matched elements}}{\text{Total number of ground truth elements}}.    
    \end{equation}

    We evaluated BBA to measure the detection accuracy of images, tables, and forms in our experiments. 
 
    \item \textbf{Caption Similarity:}  
    We evaluate caption quality using cosine similarity between the vector embeddings of the predicted ($\vec{c_p}$) and ground truth ($\vec{c_{gt}}$) captions:
    
    \begin{equation}
    \text{Similarity} = \frac{\vec{c_p} \cdot \vec{c_{gt}}}{\|\vec{c_p}\| \cdot \|\vec{c_{gt}}\|}.
    \end{equation}
    
    The embeddings are calculated using pre-trained models such as Sentence-BERT \cite{reimers2019sentence} or CLIP \cite{CLIP}.

    \item \textbf{Detection Completeness (DC):}  
    We assess detection accuracy for meaningful visual elements (e.g., tables, figures, forms) using detection completeness (DC). It is defined as the ratio of correctly identified meaningful elements to all detected elements, including non-meaningful ones (e.g., watermarks, logos, fragmented or non-visual content):
    
    \begin{equation}
        \text{DC} = \frac{\text{Number of correctly detected elements}}{\text{Total number of detected elements}}.
    \end{equation}
    
\end{itemize}



We compare our method against PDFPlumber \cite{pdfplumber}, PyMuPDF, Apache Tika \cite{tikapy,apachetika,mcnamee2012tika}, Unstructured \cite{unstructured2024}, and PDFMiner \cite{pdfminer}. As shown in Table~\ref{tab:pdf_parsing}, our pipeline consistently outperforms these baselines in visual element detection while maintaining competitive text and caption accuracy. We further compare against vision–language models, including Gemini 2.5 Pro \cite{googleGeminiAPI2025} and GPT models (GPT-4.1, GPT-4.1-mini). Results in Table~\ref{tab:comparewithVLM} show that the proposed scheme achieves better parsing accuracy than state-of-the-art VLMs with significantly lower latency. We also deployed it in our product environment due to its outstanding performance in terms of both accuracy and latency.

\subsection{Document-Based Question Answering}

For document-based QA using multimodal LLMs \cite{dong2025mmdocirbenchmarkingmultimodalretrieval,Mm-llms,Rajabzadeh23}, documents are parsed to extract text and visual elements. Visuals are rendered (as PNG/JPEG) and described by a multimodal LLM, with embeddings stored in a vector database (e.g., OpenSearch) for retrieval. At inference, the user query is combined with the top-$k$ relevant pages ($k=10$) and associated visual descriptions, then passed to a generation model to produce an answer.

\begin{table}[ht]
\centering
\begin{tabular}{|l|c|c|c|}
\hline
\textbf{Tool} & \textbf{data1} & \textbf{data2}  & \textbf{data3} \\
\hline 
\texttt{PyMuPDF}      & 4.14        & 4.12 &  3.03   \\ 
\texttt{DocLayout-YOLO} &4.20 & 4.27 & 3.18 \\
\texttt{Tika-based}      & 4.16       & 4.11   & 2.98  \\ 
\texttt{GPT4.1}      &  4.22      &  4.25  & 3.21 \\ 
\texttt{GPT4.1-mini}      &   4.13     & 4.18  & 3.10\\ 
\texttt{Llama4-Maverick}      & 4.20       &  4.19  & 3.15 \\
\texttt{Llama4-Scout}      &   4.18     &  4.24 & 3.17 \\
\texttt{xai.grok-4}      &    4.21    &   4.17 & 3.04 \\ 
\texttt{Proposed}    & \textbf{4.52} & \textbf{4.49}  & \textbf{3.52}   \\
\hline
\end{tabular}
\caption{Document question answering pipeline evaluation, with PyMuPDF, Tika+PDFBox+Unstructured, DocLayout-YOLO, GPT4.1, GPT4.1-mini, Llama4.0 Maverick, Llama4.0 Scout, xai.grok-4, and our proposed method as PDF parsers, while keeping the rest of the RAG pipeline identical. \textbf{{data1}}: our internal dataset. \textbf{data2}: MMDocRAG.  \textbf{{data3}}: PDF-VQA.}
\label{tab:qa}
\end{table}

To evaluate the quality of generated answers, we adopt an LLM-as-a-judge framework using GPT-4.1. Each predicted answer is assessed across five criteria:
\begin{enumerate}
    \item \textbf{Fluency}
    \item \textbf{Citation quality}
    \item \textbf{Text-image coherence}
    \item \textbf{Reasoning logic}
    \item \textbf{Factual accuracy}
\end{enumerate}
Each criterion is scored on a scale of 0-5, and we use the average of the five scores as the final evaluation score. 
This evaluation complements standard metrics for text extraction, visual element detection, and caption quality to comprehensively measure pipeline performance.

\begin{table}[h]
\centering
\begin{tabular}{|l|c|c|c|}
\hline
\textbf{Tool} & \textbf{data1} & \textbf{data2} & \textbf{data3}  \\
\hline   
\texttt{LLM as judge}    & \textbf{4.50}   & \textbf{4.48}  &  \textbf{3.52} \\
\texttt{Human as judge}      & 4.52       & 4.44      &  3.54 \\
\hline
\end{tabular}
\caption{Comparison of LLM-based evaluation versus human evaluation.  \textbf{{data1}}: our internal dataset. \textbf{data2}: MMDocRAG.  \textbf{{data3}}: PDF-VQA.}
\label{tab:humanvsllm}
\end{table}

We evaluated PDF parsing approaches including PyMuPDF, a hybrid Tika–PDFBox–Unstructured pipeline, DocLayout-YOLO, and VLMs (GPT-4.1, GPT-4.1-mini, LLaMA 4.0 Maverick/Scout, Grok-4 ). All pipelines rendered detected visuals as PNGs, described them with a multimodal LLM, and indexed embeddings in OpenSearch; retrieval and generation steps were identical. Experiments used MMDocRAG (220 PDFs, 14,763 pages, 2,000 questions), PDF-VQA (combination of PDFVQA, DocVQA and PDF-MVQA), and an internal dataset of 215 expert-labeled Q–A pairs. Answers were evaluated using GPT-4.1 across five criteria, with averages reported in Table~\ref{tab:qa}; statistical testing confirmed the proposed method significantly outperforms the next-best approach ($p=0.007$).

We also evaluated the reliability of the LLM judge by comparing it with human assessment. Our experts scored the answer generation results on a scale of 1 to 5 based on the same five defined criteria. 
This evaluation was performed on 50 randomly selected question-answer pairs from our internal dataset, 150 pairs from the MMDocRAG dataset, and 60 pairs from the PDFVQA dataset. 
Table~\ref{tab:humanvsllm} compares the human and LLM judgments, showing strong alignment between the two.

\subsection{Ablation Studies}

We performed ablation studies to quantify the impact of individual pipeline components: table extraction, form detection, image deduplication, artifact filtering, caption extraction, and text extraction. Each study removed one component and compared results against the full system on both our internal dataset and MMDocRAG. Table~\ref{tab:ablationstudy} shows that every component contributes to overall performance.

\begin{table}[ht]
\centering
\begin{tabular}{|l|c|c|c|}
\hline
\textbf{Component Removed} & \textbf{{data1}} & \textbf{{data2}} &  \textbf{{data3}} \\
\hline   
\texttt{Table extraction}    & 3.17   & 3.73   & 3.14 \\
\texttt{Form detection}      & 4.01       & 4.23 &   3.41  \\
\texttt{Dedup and merge}      & 4.11       & 4.08    & 3.28  \\
\texttt{Filter artifact}      & 4.19       & 4.15   & 3.39   \\
\texttt{Caption extraction}      & 3.86       & 4.04  & 3.32    \\ 
\texttt{Text extraction}      & 4.43       & 4.36 & 3.45     \\ 
\texttt{Proposed}      & \textbf{4.52}      & \textbf{4.49}  & \textbf{3.52}      \\ 
\hline
\end{tabular}
\caption{
Ablation studies evaluating the contribution of each component in our PDF parsing pipeline. 
Each row corresponds to the removal of a specific component, while the final row reports results for the complete system. Dedup and merge: Deduplication and merging of fragmented images. \textbf{{data1}}: our internal dataset. \textbf{data2}: MMDocRAG.  \textbf{{data3}}: PDF-VQA.}
\label{tab:ablationstudy}
\end{table}

Analysis of low-scoring cases revealed common failure modes: complex flowcharts, tables with arrows or intricate inter-block relationships, fine-print forms, and queries requiring information across many pages ($\geq7$). Future work will focus on reliably extracting complex visuals, interpreting interlinked table structures, and aggregating multi-page information to improve answer accuracy.

\section{Conclusions}

We present a comprehensive production level PDF parsing pipeline integrating text extraction with robust detection of images, tables, forms, and captions. By leveraging structural heuristics and visual cues, our method accurately identifies and localizes key document elements, even without embedded form metadata. Extensive evaluation on various datasets demonstrates substantial gains in extraction quality and downstream document-based QA with much lower latency. The proposed solution has been deployed in our product line.

\section{Limitations}

While the proposed framework demonstrates strong performance across diverse benchmarks and production data, several limitations remain.

\paragraph{Reliance on heuristic thresholds.}
Many components in the pipeline—such as table detection, form identification, artifact filtering, and image deduplication—rely on empirically chosen thresholds (e.g., alignment counts, size ratios, overlap ratios, frequency cutoffs). Although these thresholds generalize well across our evaluated datasets, extreme document styles or atypical layouts may require retuning. Fully eliminating such hyperparameters remains challenging without introducing heavier learned models.

\paragraph{Sensitivity to low-quality scanned documents.}
For scanned PDFs with severe noise, low resolution, or skewed layouts, the quality of text blocks extracted by the underlying PDF parser can degrade, which in turn affects downstream table detection and caption association. While our method is robust to moderate OCR noise, it does not explicitly incorporate image-based OCR correction or document de-skewing, which could further improve robustness on archival or low-quality scans.

\paragraph{Limited semantic understanding of complex visuals.}
Although semantic similarity is used to associate captions with visual elements, the framework does not perform deep visual understanding of the content within figures, charts, or diagrams (e.g., interpreting plotted values or diagram semantics). As a result, caption association may still fail for visuals whose meaning is weakly correlated with surrounding text or where captions are highly abstract or metaphorical.

\paragraph{Handling of unconventional layouts.}
Documents with highly unconventional layouts—such as overlapping figures, dense multi-column grids with interleaved captions, or visually embedded captions inside images—remain challenging. In such cases, spatial heuristics may be insufficient to disambiguate associations, particularly when multiple candidate captions fall within similar proximity ranges.




\section{Ethical Considerations}
While the proposed PDF parsing framework offers significant benefits for document understanding, search, and automation, there might be privacy considerations. The framework can process PDFs containing sensitive personal or confidential information (e.g., financial records, medical forms, legal contracts). Even when operating correctly, automated extraction of structured information increases the risk of large-scale privacy breaches. We recommend integrating robust content filtering to detect and redact sensitive data (e.g., names, addresses, ID numbers) prior to storage or sharing, and enforcing strict access controls and encryption for all intermediate outputs.

{
    \small
    \bibliographystyle{ieeenat_fullname}
    \bibliography{main}

@misc{DOCLAYOUT-YOLO,
title= {DOCLAYOUT-YOLO: ENHANCING DOCUMENT LAYOUT ANALYSIS THROUGH DIVERSE SYNTHETIC DATA
AND GLOBAL-TO-LOCAL ADAPTIVE PERCEPTION},
author={Zhiyuan Zhao and Hengrui Kang and Bin Wang and Conghui He},
doi = {https://arxiv.org/pdf/2410.12628},
year = 2024
}

@misc{Mm-llms,
title= {Mm-llms: Recent advances in multimodal large language models},
author={Duzhen Zhang and Yahan Yu and Chenxing Li and Jiahua Dong and Dan Su and Chenhui Chu and Dong Yu},
doi = {arXiv:2401.13601},
year = 2024
}

@article{pdfvqa2,
title= {PDF-MVQA: A Dataset for Multimodal Information Retrieval in PDF-based Visual Question Answering},
author={Yihao Ding and Kaixuan Ren and Jiabin Huang and Siwen Luo and Soyeon Caren Han},
journal = {arXiv:2404.12720},
year = 2024
}

@article{docvqa,
  author={Mathew, M. and Karatzas, D. and Jawahar, C.V.}, 
  title={Docvqa: A dataset for vqa on document
images},
  year={2021},
journal = {Proceedings of the IEEE/CVF Winter Conference on Applications
of Computer Vision (WACV)}   
}

@inproceedings{optuna,
  title={{O}ptuna: A Next-Generation Hyperparameter Optimization Framework},
  author={Akiba, Takuya and Sano, Shotaro and Yanase, Toshihiko and Ohta, Takeru and Koyama, Masanori},
  booktitle={The 25th ACM SIGKDD International Conference on Knowledge Discovery \& Data Mining},
  pages={2623--2631},
  year={2019}
}

@article{pdfvqa,
  author={Ding, Yihao
          and Luo, Siwen
          and Chung, Hyunsuk
          and Han, Soyeon Caren}, 
  title={PDF-VQA: A New Dataset for Real-World VQA on PDF Documents},
  booktitle="Machine Learning and Knowledge Discovery in Databases: Applied Data Science and Demo Track",
  year={2023},
journal = {European Conference on Machine Learning and Principles and Practice of Knowledge Discovery in Databases}   
}

@article{pdfmvqa,
  author={Yihao Ding and Kaixuan Ren and Jiabin Huang and Siwen Luo and Soyeon Caren Han}, 
  title={PDF-MVQA: A Dataset for Multimodal Information Retrieval in PDF-based Visual Question Answering},
  year={2024},
  journal={arXiv:2404.12720v1}   
}

@misc{googleGeminiAPI2025,
  title = {Document understanding | Gemini API | Google AI for Developers},
  author = {{Google AI}},
  year = {2025},
  note = {Accessed: 2025-07-09},
  url = {https://ai.google.dev/gemini-api/docs/document-processing}
}

@inproceedings{zhong2019publaynet,
  author    = {Xu Zhong and Jianbin Tang and Antonio Jimeno Yepes},
  title     = {{PubLayNet: Largest Dataset Ever for Document Layout Analysis}},
  booktitle = {2019 International Conference on Document Analysis and Recognition (ICDAR)},
  year      = {2019},
  pages     = {1015--1022},
  doi       = {10.1109/ICDAR.2019.00165}
}

@inproceedings{li2020docbank,
  author    = {Yiheng Li and Lei Cui and Shaohan Huang and Furu Wei and Ming Zhou and Zhoujun Li},
  title     = {{DocBank: A Benchmark Dataset for Document Layout Analysis}},
  booktitle = {Proceedings of the 28th International Conference on Computational Linguistics (COLING)},
  year      = {2020},
  pages     = {949--960},
  url       = {https://aclanthology.org/2020.coling-main.82}
}

@misc{pdf32000,
  title={PDF 32000-1:2008 -- Document Management -- Portable Document Format -- Part 1: PDF 1.7},
  author={{Adobe Systems Incorporated}},
  year={2008},
  note={ISO 32000-1:2008},
  url={https://www.adobe.com/content/dam/acom/en/devnet/pdf/pdfs/PDF32000_2008.pdf}
}

@misc{pymupdf_docs,
  author = {{Artifex Software Inc.}},
  title = {{PyMuPDF Documentation}},
  year = {2023},
  note = {Accessed: 2025-07},
  url = {https://pymupdf.readthedocs.io/}
}

@misc{tika,
  author    = {Apache Software Foundation},
  title     = {Apache Tika: A Content Analysis Toolkit},
  howpublished = {\url{https://tika.apache.org/}},
  year      = {2024},
  note      = {Accessed: 2024-06-24}
}

@misc{unstructured,
  author    = {Unstructured Technologies and contributors},
  title     = {unstructured: An Open-Source Toolkit for Document Parsing},
  howpublished = {\url{https://github.com/Unstructured-IO/unstructured}},
  year      = {2024},
  note      = {Accessed: 2024-06-24}
}

@misc{Rajabzadeh23,
  author    = {Hossein Rajabzadeh and Suyuchen Wang and Hyock Ju Kwon and Bang Liu},
  title     = {Multimodal multi-hop
question answering through a conversation between tools and efficiently finetuned large language
models},
  year      = {2023},
  doi      = {arXiv:2309.08922}
}

@misc{dong2025mmdocirbenchmarkingmultimodalretrieval,
      title={MMDocIR: Benchmarking Multi-Modal Retrieval for Long Documents}, 
      author={Kuicai Dong and Yujing Chang and Xin Deik Goh and Dexun Li and Ruiming Tang and Yong Liu},
      year={2025},
      eprint={2501.08828},
      archivePrefix={arXiv},
      primaryClass={cs.IR},
      url={https://arxiv.org/abs/2501.08828}, 
}

@misc{dong2025benchmarkingretrievalaugmentedmultimomalgeneration,
      title={Benchmarking Retrieval-Augmented Multimomal Generation for Document Question Answering}, 
      author={Kuicai Dong and Yujing Chang and Shijie Huang and Yasheng Wang and Ruiming Tang and Yong Liu},
      year={2025},
      eprint={2505.16470},
      archivePrefix={arXiv},
      primaryClass={cs.IR},
      url={https://arxiv.org/abs/2505.16470}, 
}

@misc{pdfminer,
  author       = {Yusuke Shinyama},
  title        = {PDFMiner: Python PDF parser and analyzer},
  year         = {2014},
  howpublished = {\url{https://github.com/euske/pdfminer}},
  note         = {Accessed: 2025-07-15}
}

@inproceedings{reimers2019sentence,
  title     = {Sentence-BERT: Sentence Embeddings using Siamese BERT-Networks},
  author    = {Reimers, Nils and Gurevych, Iryna},
  booktitle = {Proceedings of the 2019 Conference on Empirical Methods in Natural Language Processing (EMNLP)},
  year      = {2019},
  pages     = {3982--3992},
  publisher = {Association for Computational Linguistics},
  url       = {https://aclanthology.org/D19-1410},
  doi       = {10.18653/v1/D19-1410}
}

@article{levenshtein1966binary,
  title     = {Binary Codes Capable of Correcting Deletions, Insertions, and Reversals},
  author    = {Levenshtein, Vladimir I.},
  journal   = {Soviet Physics Doklady},
  volume    = {10},
  number    = {8},
  pages     = {707--710},
  year      = {1966},
  publisher = {American Institute of Physics}
}

@inproceedings{CLIP,
  title     = {Learning Transferable Visual Models From Natural Language Supervision},
  author    = {Radford, Alec and Kim, Jong Wook and Hallacy, Luke and Ramesh, Aditya and Goh, Gabriel and Agarwal, Sandhini and Sastry, Girish and Askell, Amanda and Mishkin, Pam and Clark, Jack and Krueger, Gretchen and Sutskever, Ilya},
  booktitle = {Proceedings of the 38th International Conference on Machine Learning (ICML)},
  year      = {2021},
  publisher = {PMLR},
  series    = {Proceedings of Machine Learning Research},
  volume    = {139},
  pages     = {8748--8763},
  url       = {https://proceedings.mlr.press/v139/radford21a.html}
}

@article{liu2013,
    author  = "Kefei Liu and João Paulo C. L. da Costa and Hing Cheung So and ‪André L. F. de Almeida",
    title   = "Semi-blind receivers for joint symbol and channel estimation in space-time-frequency MIMO-OFDM systems",
    year    = "2013",
    journal = "IEEE Transactions on Signal Processing"
}

@article{liu2016,
    author  = "Kefei Liu and João Paulo C. L. da Costa and Hing Cheung So and Lei Huang and Jieping Ye",
    title   = "Detection of number of components in CANDECOMP/PARAFAC models via minimum description length",
    year    = "2016",
    journal = "Digital Signal Processing"
}

@inproceedings{Wang2020Logo2K,
author={Jing Wang and and Weiqing Min and and Sujuan Hou and Shengnan Ma and Yuanjie Zheng and Haishuai Wang and Shuqiang Jiang},
booktitle={AAAI Conference on Artificial Intelligence. Accepted},
title={{Logo-2K+:} A Large-Scale Logo Dataset for Scalable Logo Classification},
year={2020}
}

@misc{pymupdf,
  author       = {{Artifex Software Inc.}},
  title        = {{PyMuPDF: Python bindings for MuPDF}},
  year         = {2024},
  howpublished = {\url{https://pymupdf.readthedocs.io/}},
  note         = {Version 1.23.22}
}

@misc{pdfplumber,
  author       = {Jeremy Singer-Vine},
  title        = {{pdfplumber: Python library for extracting information from PDFs}},
  year         = {2024},
  howpublished = {\url{https://github.com/jsvine/pdfplumber}},
  note         = {Version 0.10.3}
}

@book{mcnamee2012tika,
  title     = {Tika in Action},
  author    = {Chris A. Mattmann and Jukka Zitting},
  year      = {2012},
  publisher = {Manning Publications Co.},
  isbn      = {9781617290084}
}

@misc{apachetika,
  title        = {Apache Tika},
  author       = {{Apache Software Foundation}},
  howpublished = {\url{https://tika.apache.org/}},
  year         = {2025}
}

@misc{tikapy,
  author = {Mattmann, Chris A.},
  title = {tika-python: Python bindings to Apache Tika},
  year = {2024},
  howpublished = {\url{https://github.com/chrismattmann/tika-python}},
  note = {Version 1.24}
}

@misc{unstructured2024,
  author       = {Unstructured Technologies},
  title        = {unstructured: A library for preprocessing and parsing unstructured data},
  year         = {2024},
  howpublished = {\url{https://github.com/Unstructured-IO/unstructured}},
  note         = {Version 0.11.0}
}
}


\end{document}